\documentclass[10pt,authoryear,final]{elsarticle}

\usepackage{graphicx} 
\graphicspath{{./figures/}} 
\usepackage{amsmath} 
\usepackage{amssymb} 
\usepackage{pbox}
\usepackage{physics}
\usepackage{dsfont}
\usepackage{siunitx}
\usepackage{color}
\usepackage{xcolor}
\usepackage{soul}
\usepackage{natbib}
\usepackage{orcidlink}
\usepackage{hyperref}
\usepackage[final]{changes}

\begin{document}
\begin{frontmatter}
  \title{Scaling up machine learning-based chemical plant simulation: A method for fine-tuning a model to induce stable fixed points}
  \author[affiliation_TU,affiliation_BASLEARN]{Malte Esders\corref{cor} \orcidlink{0000-0002-9136-914X}}
  \ead{esders@tu-berlin.de}
  \author[affiliation_BASF,affiliation_INEOS]{Gimmy Alex Fernandez Ramirez \orcidlink{0009-0005-2636-1175}}
  \ead{gimmy.fernandez@ineos.com}
  \author[affiliation_TU,affiliation_BASLEARN]{Michael Gastegger \orcidlink{0000-0001-7954-3275}}
  \ead{michael.gastegger@tu-berlin.de}
  \author[affiliation_BASF]{Satya Swarup Samal\corref{cor} \orcidlink{0000-0001-8909-7168}}
  \ead{satya-swarup.samal@basf.com}
  \cortext[cor]{Corresponding author}
  \address[affiliation_TU]{Machine Learning Group, Technische Universität Berlin, 10587 Berlin, Germany}
  \address[affiliation_BASLEARN]{BASLEARN – TU Berlin/BASF Joint Lab for Machine Learning, Technische Universität Berlin, 10587 Berlin, Germany}
  \address[affiliation_BASF]{BASF SE, 67056 Ludwigshafen, Germany}
  \address[affiliation_INEOS]{INEOS Styrolution Group GmbH, 67056 Ludwigshafen, Germany}

\begin{abstract}
  Idealized first-principles models of chemical plants can be inaccurate. An alternative is to fit a Machine Learning (ML) model directly to plant sensor data. We use a structured approach: Each unit within the plant gets represented by one ML model. After fitting the models to the data, the models are connected into a flowsheet-like directed graph. We find that for smaller plants, this approach works well, but for larger plants, the complex dynamics arising from large and nested cycles in the flowsheet lead to instabilities in the  solver during model initialization. We  show that a high accuracy of the single-unit models is not enough: The gradient can point in unexpected directions, which prevents the solver from converging to the correct stationary state. To address this problem, we present a way to fine-tune ML models such that initialization, even with very simple solvers, becomes robust.
\end{abstract}

\end{frontmatter}

\section{Introduction}
\begin{figure}[h]
  \centering
  \includegraphics[width=0.99\linewidth]{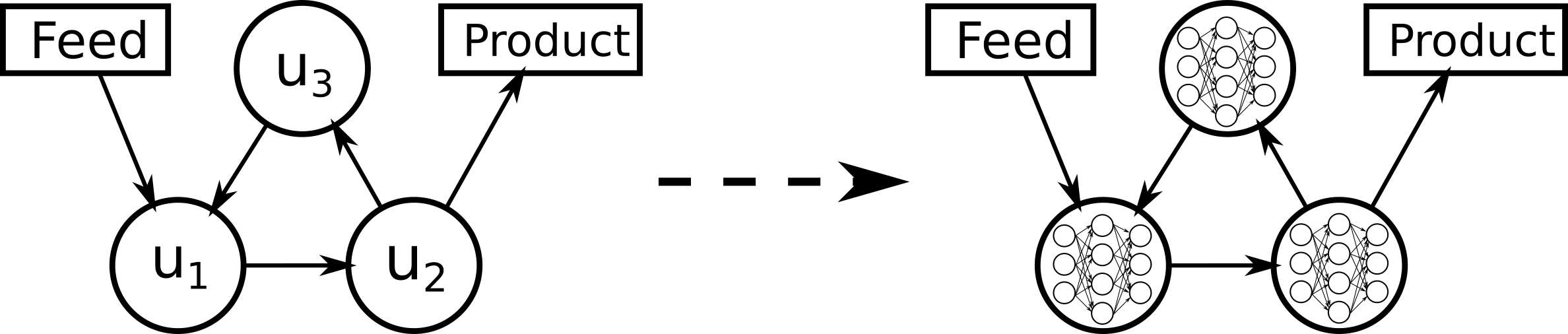}
  \caption{Left: Flowsheet schematic. $u_1$, $u_2$ and $u_3$ are process units. The arrows indicate in which direction the chemicals flow. Right: In the present study, the first-principles-based simulation for each process unit is turned into an ML model, in this case, NNs.}
  \label{fig:structnet_schematic}
\end{figure}

Simulation is the default tool for planning and operating chemical plants. For a number of reasons outlined below, idealized simulations can give inaccurate results compared to the actual, brick-and-mortar plant.  To achieve more accurate results, creating a model directly from sensor data is enticing. Machine Learning (ML) is ideally suited for such a data-based modeling task. We applied the conventional structural ML modeling approach to a larger plant and found that during model initialization, solvers fail to converge or converge towards a stationary point that is not equal to the true steady-state of the plant. In this paper, we show that this problem happens due to the presence of large and nested cycles within the plant, during which the gradient direction becomes infeasible when using ML models. We introduce a method to fine-tune an ML model without using additional data in such a way that the model gradients enable solvers find the true stationary solution robustly. The trick we employ is to mimick the model initialization procedure already during the model identification stage. Using an end-to-end loss gradient obtained from this procedure lets us fine-tune the single-unit models in such a way that their accuracy barely changes, but their gradient changes towards a useful direction.\\
Data-driven models are needed in the first place because the first-principles simulation sometimes is not a good approximation of the real plant. Reasons for this include that some units of the plant may be too complex to be captured with equations, either because the exact equations are not known or because the full set of equations would be too large for a solver to converge. Another limitation is that existing simulation software cannot mathematically express each piece of equipment to the exact specifications. Additionally, time-based changes within the plant make the plant operation increasingly differ from the simulation \citep{bogojeski2021forecasting,gordon2022data}.\deleted{ Examples of such changes are catalyst degradation, the fouling of a heat exchanger, or the accumulation of byproducts.}\\
\replaced{We use structured models, which means that we}{Chemical plants can become highly complex, but this complexity arises from connecting a collection of much simpler units. We utilize this structure and} model each process unit individually, as opposed to modeling the entire plant as a black box\deleted{. Specifically, we model each process unit with a neural network (NN) and connect all NN models into an overall model representing the structure of the underlying flowsheet} (see fig.~\ref{fig:structnet_schematic} for an illustration). The advantage of a structured  model is having much finer-grained control over the internal processes. For instance, it becomes possible to answer questions like "What would happen if we increase the pressure after unit XYZ?". On the other hand, modeling each unit individually also introduces a challenge later on during model initialization if the flowsheet contains large or nested cycles.\\
\deleted{Cycles in the flowsheet-graph are standard in chemical plants, e.g., due to recycling material from later stages in the process being fed back into the beginning of the plant or when using a heat exchanger. A nested cycle is present if at least one unit is part of two cycles simultaneously.\\}
There are two approaches of solving flowsheets with cycles: \textit{equation oriented} and \textit{sequential modular}, or a hybrid of the two \citep{byrne2000global,bongartz2019deterministic}. The more common one is \textit{equation oriented} solving \citep{shacham1982equation}, where the equations governing each piece of equipment are solved simultaneously.  ML-based structured models are typically solved in the \textit{sequential modular} approach \citep{bubel2021modular,zapf2021gray}\added{, where each piece of equipment is solved one after the other, which can be simpler but requires special treatment of cycles}. \deleted{In the sequential modular approach, one starts at the inputs to the flowsheet and computes the output of the first process unit. This output becomes the input to the next process unit, and so on. A problem is encountered at the unit where a cycle stream terminates (fig.~\ref{fig:tear_stream}): values for the variables of the cycle stream are not known, therefore computing the output of that unit is not possible (see section \ref{sec:cycle_solving} for a more thorough explanation). An initial guess for the cycle stream has to be used, and this initial guess is then refined iteratively. }A plethora of cycle-solving algorithms have been published \citep{wegstein1958accelerating,orbach1971convergence,crowe1975convergence}, however\deleted{, as we show here}, none of the four solve methods we tested worked well for solving ML-based cycles. \\

Applications of ML models to chemical process engineering have been explored for several decades. The first studies comparable to ours were published by \citet{palmer2002metamodeling,palmer2002optimization}, who replaced units in the flowsheet simulation with polynomial and kriging models and performed optimization of the cost of operation. Recently, there has been a resurgence of interest in this topic (for an overview of recent progress see \citet{lee2018machine,schweidtmann2022optimization}). ML models have been combined with physics-based equations for increased accuracy \citep{bikmukhametov2020combining}, and have been used to optimize various units \citep{shalaby2021machine,briceno2023machine} and entire flowsheets \citep{caballero2008algorithm,burre2022global}. The current study deals with the problem of model initialization in purely ML-based models. Finding a solution for model initialization can be a tedious and difficult task, which is often ommitted from resulting publications \citep{casella2021choice}. ML models have been used to explore the solution space of flowsheet simulations \citep{heese2019optimized,zapf2021gray,ludl2022using}. The current study is different in that the goal is not to create a classical gray-box surrogate model. Rather, the goal is to develop an approach with which one could fit a structured model to an existing plant. This significantly constrains the type of data that can be used: Surrogate modeling studies typically simulate each process unit separately and generate one dataset per unit, and then fit each single-unit model with its own dataset \citep{bubel2021modular}. In this study, an entire plant is simulated at once to create one overall dataset, from which each single-unit ML model is fitted. The difference is that when simulating an entire plant, the admissible solutions are severely restricted compared to simulating each unit separately, which leads to less variable data. When fitting ML models on less variable datasets, they become worse at generalization. This unfortunate limitation has to be confronted when moving towards real-world data. To the best of our knowledge, the current study is the first purely ML-based modeling approach using this kind of whole-plant data. \\
In this work, we start by showing that the presence of large and nested cycles leads to instabilities in all tested solve methods. We stress that these instabilities are not a niche problem of our particular example plant: Instead, such problems have to be expected whenever a plant contains large or nested cycles, as is common in chemical plants. In the next step, we propose a fine-tuning method to fix this problem. In this fine-tuning step, the same data as before is used, i.e. no new data is required. We show qualitatively and quantitatively that after applying our proposed fine-tuning method, the solving of cycles becomes robust and trivial. \\
Our contributions in this study are as follows:
\begin{enumerate}
  \item We scale up existing approaches for structured modeling to significantly more complex processes. We provide a qualitative and quantitative analysis of existing approaches at this scale.
  \item We show that without modifications, established methods fail to produce single-unit models which are suitable for solving large or nested cycles during model initialization. We show that the reason \deleted{for this} is that the gradient along large or nested cycles is not constrained and can point in arbitrary directions.
  \item We introduce a method that fine-tunes NN  models such that even extremely simple solvers converge towards the true stationary state.
\end{enumerate}

\section{Methods}
\subsection{Synthetic model of Cumene process}
\label{sec:cumene_process}
\begin{figure*}[h]  
  \centering
  \includegraphics[width=1.0\linewidth]{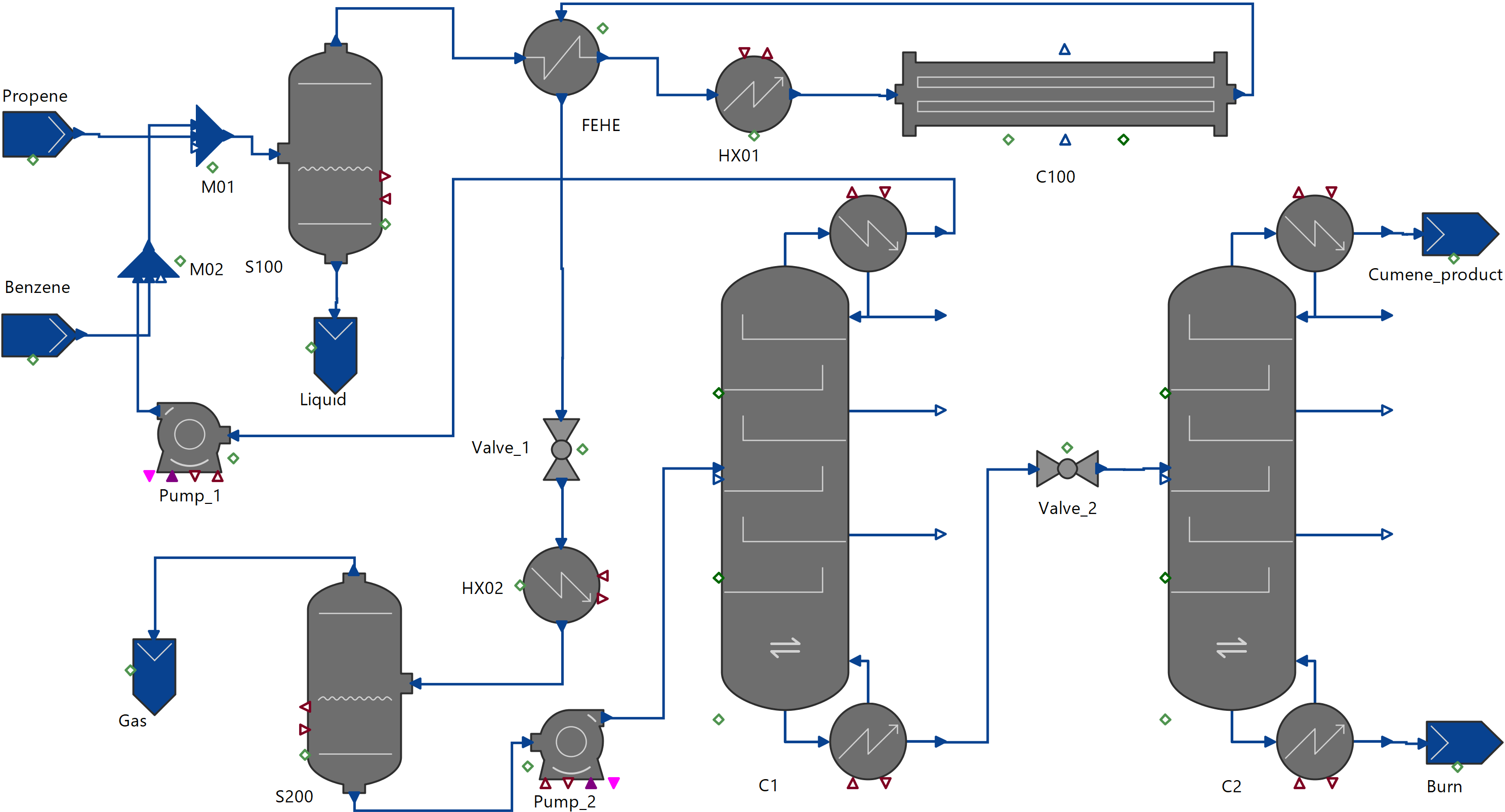}
  \caption{Cumene process flowsheet. For the type of each unit, refer to Table \ref{table:single_unit_errors}}
  \label{fig:cumene_process}
\end{figure*}
As the basis for our experiments, we use the full Cumene process, which is a well-studied and complex process. The process model corresponds to the same used by Luyben \citep{luyben2010design}; kinetic data were taken from the same reference. Thermodynamic properties were estimated using the Redlich-Kwong-Suave equation of state provided by Multiflash (Multiflash Command Reference, Infochem / KBC Advanced Technologies plc., Version 4.4, page 39) included in process simulation package gPROMS (Siemens Process Systems Enterprise Ltd. London, UK). Fig.~\ref{fig:cumene_process} shows the flowsheet of the process. \\
A liquid mixture of 95 mol \% propylene and 5 mol \% of propane is mixed with a fresh stream of benzene and the recycled benzene coming from the first distillation column C1. The mixture is then fed to the S100 (a total evaporator) where it is evaporated, leaving S100 at 209°C and 25 bar. The vapors are then preheated using two heat exchangers: the first one, a feed-effluent heat exchanger (FEHE), uses part of the heat of the reactor’s effluent, the second, HX01, provides the additional heat to reach 360°C, which is the same temperature of the cooling medium of the reactor C100. C100 is a tubular reactor with 342 tubes, with 76.3 mm inside diameter, 2 mm wall thickness, and 6 m in length. It is filled with a solid catalyst of density 2000 kg/m with a void fraction of 0.5 m\textsuperscript{3}/m\textsuperscript{3}. An overall heat transfer coefficient of the cooling of 65 W/m\textsuperscript{2}K was used. \\
After the effluent is used to pre-heat the reactor's feed, it is passed through Valve1 to reduce its pressure to 1.75 bar and cooled down to 90°C. The stream is now a liquid-vapor mixture that is separated in the flash tank S200. The liquid is then fed to the column C1. The column C1 has 15 stages; the feed is located in stage 8, and the average column pressure was specified to 1.75 bar, with a reflux ratio of 0.44 mol/mol. The purity of the Cumene at the bottom was specified to be 0,0005 mol/mol. The top effluent of the column, which is rich in benzene, is recycled. The Cumene-rich effluent of C1 is fed to the twelfth stage of the column C2. C2 has 20 stages and operates at an average pressure of 1 bar. Cumene was specified to be at most 0.001 mol/mol at the bottom, with a reflux ratio of 0.63 mol/mol.

\subsection{Data generation procedure}
\label{sec:data_generation}
In order to generate data to fit the ML models, the Cumene process described in section \ref{sec:cumene_process} was simulated in various configurations. The simulation was run until it converged towards a stationarystate for various process conditions. The following variables were captured in each stream after a stationary state was reached: the total mass flowrate [\si{kg/s}], temperature [\si{K}], pressure [\si{bar}] and the chemical composition fractions [\si{kg/kg}]. The five chemicals involved are Benzene, Cumene, Diisopropylbenzene, Propane, and Propylene. All variables were measured both for the liquid and vapor phase (as long as each phase exists), such that in each stream, there were a maximum of 16 variables (otherwise 8 for just one phase). \\
For some types of equipment, there were additional variables: some types of equipment heat up the stream, in which case the heating power was captured as well [\si{kW}]. The distillation columns had the specified reflux ratio and reboiling ratio [\si{kg/kg}] as additional inputs. Additionally, the output stream of the reactor C100 included the conversion of both Benzene and Propylene [\si{kg/kg}]. The data were generated by varying one variable of a small set of variables at a time and letting the simulation find appropriate values for all other variables. \\
Table \ref{table:data_ranges} indicates the variables which were varied. We intentionally generated very few data points to best mimic real-world scenarios in which available data are typically limited. In total, 397 data points were created, of which 357 were used for fitting (i.e., model identification) and 40 were used for testing.
\begin{table}
  \begin{tabular}{lrr}
    Variable                     &  from   &  to \\
    Benzene inlet flow [kg/s]    &  1.18   &  2.68 \\
    Propene inlet flow [kg/s]    &  1.08   &  1.45 \\
    S100 output temperature [K]  &  440    &  512 \\
    HX01 output temperature [K]  &  590    &  650 \\
    Valve1 output pressure [bar] &  1      &  3 \\
    HX02 output temperature [K]  &  353    & 373 \\
  \end{tabular}
  \caption{Ranges of variables that were varied to generate the data.}
  \label{table:data_ranges}
\end{table}

\subsection{Cycle solving}
\label{sec:cycle_solving}
\begin{figure}[h]
  \centering
  \includegraphics[width=0.45\linewidth]{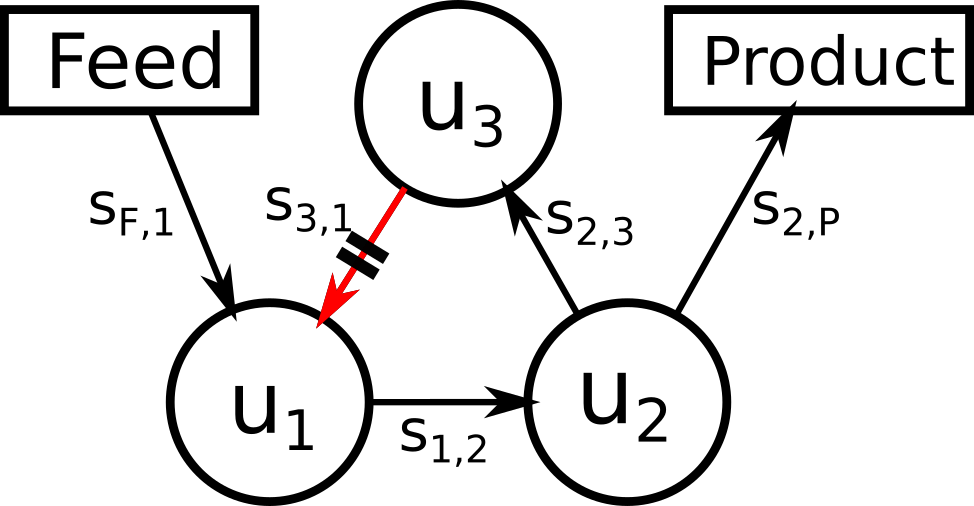}
  \caption{Flowsheet schematic with the tear stream indicated in red.}
  \label{fig:tear_stream}
\end{figure}
\begin{figure*}[h]
  \centering
  \includegraphics[width=0.99\linewidth]{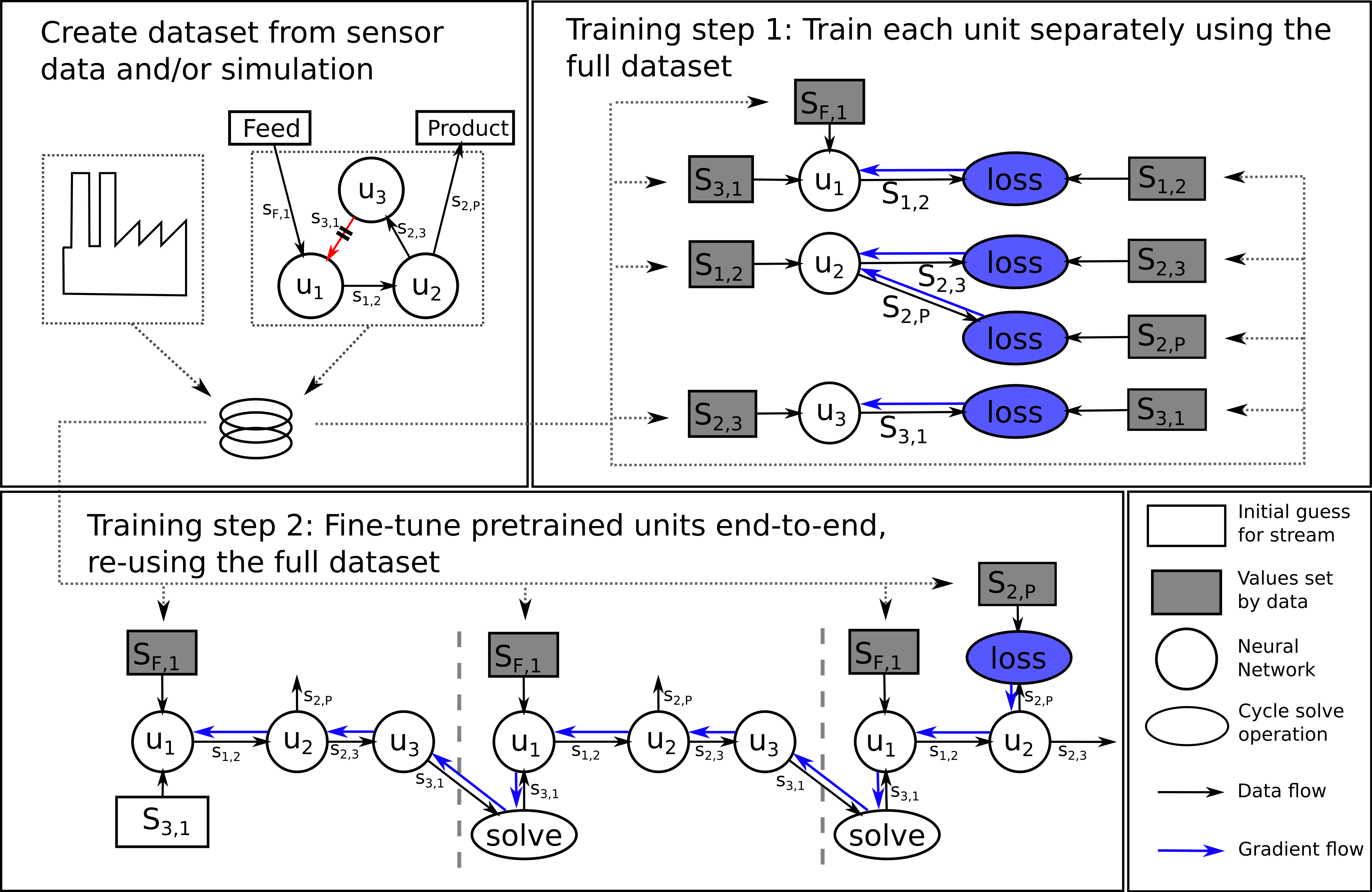}
  \caption{Top left: Data are generated either from sensors in a real plant, or from a first-principles-based simulation. Top right: The ML models representing each flowsheet unit are fitted entirely isolated from each other, with each input/output set by the collected data. Bottom: The flowsheet is sequentially predicted from beginning to end, including the iterative cycle-solving (forward pass, black arrows). Afterward, the gradient is backpropagated through each node along the unrolled forward pass (blue arrows).}
  \label{fig:step_by_step}
\end{figure*}
When propagating information between the nodes in the flowsheet, from the flowsheet input (Feed) to the output (Product), we assume that the only information given is the Feed. All other intermediate streams are considered unknown and must be predicted sequentially. In the presence of cycles however, this sequential procedure encounters a problem at the first unit where a cycle in the flowsheet terminates: As an example, take unit $u_1$ (fig.~\ref{fig:tear_stream}). Let $u_1(s_{F,1}, s_{3,1})$ also be the symbol for the function computing the output of unit $u_1$. While the input coming from Feed in stream $s_{F,1}$ is known, the stream from $u_3$, $s_{3,1}$, is not. Computing $u_1(s_{F,1}, s_{3,1})$ is therefore impossible. \\
In order to find values for the streams along the cycle, an iterative cycle-solving procedure must be employed. The first step is to identify a \textit{tear stream}. Any stream along a cycle can be chosen to be the tear stream. For the purpose of cycle solving, this tear stream is considered to be cut, thus breaking the cycle. An initial guess for the tear stream values is made, and with it, the rest of the streams along the cycle are predicted sequentially (black continuous arrows in the bottom part of fig.~\ref{fig:step_by_step}). Iteratively, the initial guess for the tear stream is improved every time the sequential prediction reaches the tear stream. Selecting a stream to be the tear stream is possible by visual inspection (by human designers) for simple flowsheets. For more intricate flowsheets, tear stream-finding algorithms exist \citep{pho1973tear}. \\
It is important to note that we consider the steady-state operation of a chemical plant. This means that all streams stay constant over time. When finding values for the tear stream, one needs to make sure that the obtained values are solutions of such a stationary state. This means that when using the sequential modular approach to finding values for the tear stream, the estimate of the tear stream from one iteration needs to be equal to the next.\\
Mathematically, the iterative solving of a flowsheet cycle corresponds to fixed point iteration. Fixed point iteration is the repeated application of a function $f$ to an initial value $x_0$.
\begin{equation}
  \begin{aligned}
    x_1 & = f(x_0) \\
    x_{k+1} & = f(x_k) \\
  \end{aligned}
\end{equation}
Under certain conditions, the obtained sequence $x_0, x_1, ..., x_k$ is converging towards a fixed point. A fixed point is found if for some $k$ the function value $f(x^*)$ is equal to $x^*$ itself:
\begin{equation}
  \begin{aligned}
    x^* & = f(x^*)
  \end{aligned}
  \label{eq:fixed_point_condition}
\end{equation}
To see that solving a cycle in a flowsheet is equivalent to fixed-point iteration, we consider the cycle in fig.~\ref{fig:tear_stream}. In order to simplify the notation, we ignore the streams to and from the flowsheet, $s_{F,1}$ and $s_{2,F}$. Applying the units $u_1$, $u_2$ and $u_3$ to an initial guess for the stream $s_{{3,1}_0}$, i.e., iteratively applying all units along the cycle to the cycle input, can be seen as a function composition:
\begin{equation}
  f = u_3 \circ u_2 \circ u_1
  \label{eq:flowsheet_response}
\end{equation}
The function $f$ is called the \textit{flowsheet response function}. A flowsheet response function for an example cycle is displayed in  fig.~\ref{fig:flowsheet_response}.\\
Repeated iteration through the cycle is equivalent to repeated application of the flowsheet response function and is, therefore, a fixed point iteration:
\begin{equation}
  \begin{aligned}
    s_{{3,1}_{k+1}} & = u_3 \circ u_2 \circ u_1(s_{{3,1}_{k}})\\
    s_{{3,1}_{k+1}} & = f(s_{{3,1}_{k}})
  \end{aligned}
\end{equation}

\subsubsection{Cycle solving methods used in this study}
\label{sec:cycle_solving_methods}
\begin{figure}[h]
  \centering
  \includegraphics[width=0.99\linewidth]{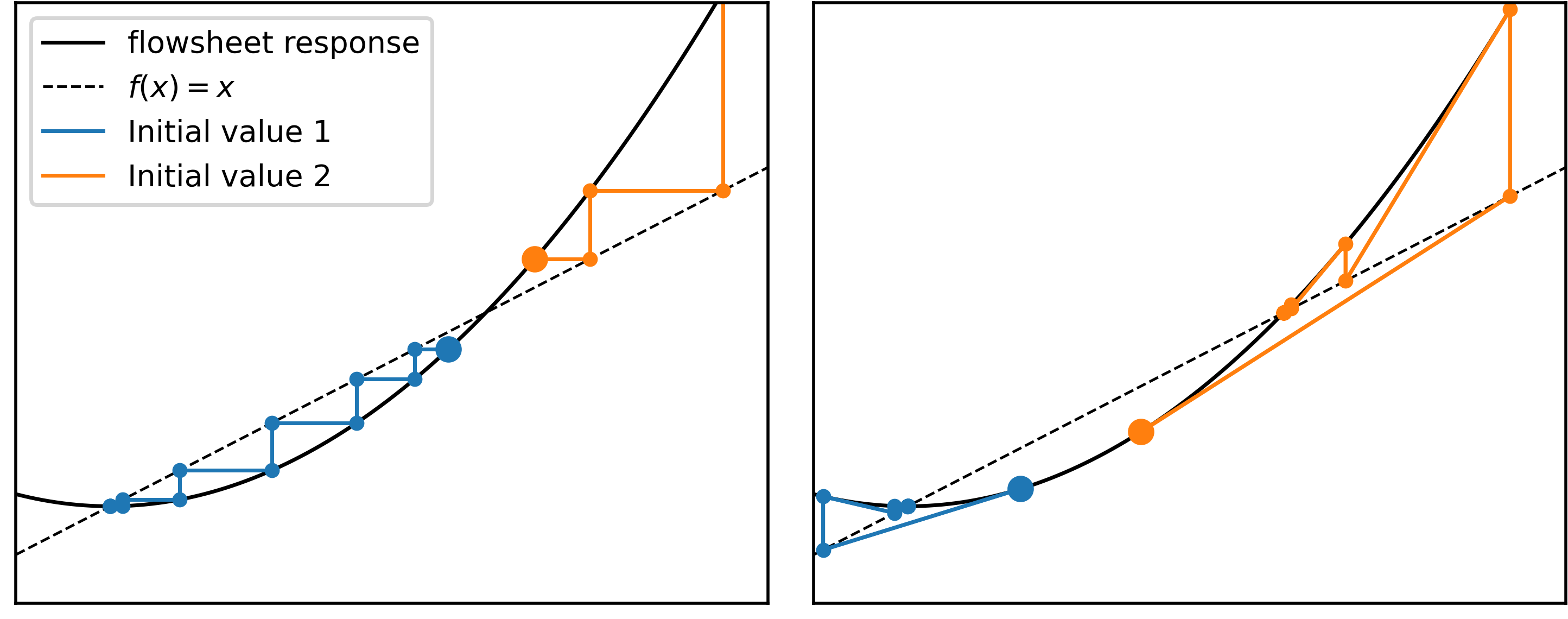}
  \caption{Visualization of a flowsheet response function and corresponding solve iterations of two different solve methods. The intersections of the flowsheet response curve (solid line) and the diagonal line where $f(x)=x$ (dotted line) are the fixed points of the flowsheet response function. Left: Direct substitution method. Right: Newton method. The blue and orange lines represent iterations for different initial values. The initial values were chosen for illustrative purposes. Each solve iteration starts at the large circles.}
  \label{fig:flowsheet_response}
\end{figure}

We compare four ways of solving nested cycles. The first method is the well-known and obvious \textit{direct substitution} method \citep{smith2005chemical}, which works simply by using the flowsheet response for the tear stream as the new guess for the next iteration:
\begin{equation}
  x_{k+1} = f(x_k)  \tag{direct substitution}
\end{equation}
Applying the direct substitution method is visualized in fig.~\ref{fig:flowsheet_response} (left). Each successive new estimate looks like a jump to the diagonal line where $f(x_k)=x_k$, resulting in the characteristic "staircase" plot. Note that for the initial value in the orange line, the iteration diverges towards infinity. In general, the direct substitution method, which is a fixed-point iteration, can only be guaranteed to converge to a unique fixed point if $f$ is a "contraction mapping", which is equivalent to saying that $f$ is Lipschitz continuous with Lipschitz constant $K < 1$ (Banach fixed-point theorem \citep{banach1922operations}). For a differentiable function $f: \mathbb{R} \rightarrow \mathbb{R}$, such a condition (contraction mapping) is fulfilled if $\sup |f'(x)| < 1$. Despite this, the direct substitution method proved to be a surprisingly robust method for smaller cycles in our experiments.\\
\\
The second method we compare is the \textit{Wegstein} method \citep{wegstein1958accelerating}. It is similarly well-known in the chemical process engineering literature \citep{smith2005chemical} and is a method that can find a solution even in a locally divergent sequence. Also, like the direct substitution method, it does not need gradient information, although it uses an approximation to the gradient:
\begin{equation}
  a = \frac{f(x_k) - f(x_{k-1})}{x_k - x_{k-1}}
\end{equation}
Using $a$, the update in the Wegstein method is:
\begin{equation}
  x_{k+1} = \frac{a}{a-1}x_k - \frac{1}{a-1}f(x_k)  \tag{Wegstein}
\end{equation}
\\
The third and fourth methods we compare are gradient-based methods. An advantage of using NNs is that the gradient is easily available without numerical approximation, which enables us to use these methods. The arguably most obvious gradient-based method is the \textit{Newton} root finding method. It can be seen that cycle solving is equivalent to root finding by a simple transformation of equation \ref{eq:fixed_point_condition}:
\begin{equation}
    0 = f(x^*) - x^*
\end{equation}
Therefore we are searching for the roots of a function $F(x) = f(x) - x$.\\
In the Newton method, the iteration towards the root is given by
\begin{equation}
  x_{k+1} = x_k - \frac{f(x_k)}{f'(x_k)}  \tag{Newton}
\end{equation}
The resulting solve iteration of the Newton method is visualized in fig.~\ref{fig:flowsheet_response} (right). If the gradient $f'(x_k)$ is close to 1, the Newton method can select points far away for $x_{k+1}$ (orange line), but in general it finds fixed points much faster than direct substitution.
\\
The fourth method is the Broyden–Fletcher–Goldfarb–Shanno algorithm (\textit{BFGS}) \citep{fletcher2013practical}. The BFGS algorithm uses an approximation to the Hessian matrix $B_k$ for finding the search direction $p_k$ along which the next estimate $x_{k+1}$ is searched. The search direction is the solution of $B_k p_k = - f'(x_k)$. After finding an appropriate step size $\alpha_k$ along $p_k$, the update in the BFGS method is
\begin{equation}
  x_{k+1} = x_k + \alpha_k p_k  \tag{BFGS}
\end{equation}

\subsection{Training step 1: Single-unit neural network training}
The method presented in this study involves two training steps (fig.~\ref{fig:step_by_step}). In the first training step, each NN was trained on its own. In the second training step, the entire flowsheet was fine-tuned end-to-end (see section \ref{sec:methods:finetuning}). For the first training step, the data for the input and output streams belonging to the unit represented by a given NN were extracted from the data, and the NN was trained in the typical ML fashion to predict the output based on the input. \\
The NN architecture for each single-unit model was two hidden layers with 100 neurons each. A skip connection in the form of a dense layer without an activation function from the input to the output was used. The activation function of the hidden layers was the \textit{Softplus} function. All in all the structured model including all single-unit models contained $178,000$ parameters. Before forwarding through a NN, the input was normalized by subtracting the data mean and dividing by the standard deviation. The output of each NN was reverse-normalized again with the mean and standard deviation of the output stream. This enabled the NN to work with and predict values in reasonable data ranges and proved crucial for obtaining a good performance. For more training details, see \ref{sec:training_details}.

\subsection{Training step 2: fine-tuning through the cycle solver}
\label{sec:methods:finetuning}
During inference, while using a solver to find solutions for the cycles in the flowsheet, the partial models are exposed to inputs that are not (necessarily) a solution of the stationary state. Consider unit $u_1$ in fig.~\ref{fig:tear_stream}. The partial model for unit $u_1$ receives input in stream $s_{F,1}$, and in stream $s_{3,1}$ it receives the initial guess for the tear stream (typically the training dataset mean). Since the initial guess is not related to the new input, this combination of new input and initial guess is (typically) not a solution of the stationary state, and therefore the model for $u_1$ has not been exposed to such combined input in its fitting procedure in training step 1. Moreover, the gradients of the variables in the tear stream point in diverging directions and prevent the solver from converging (see section \ref{sec:results:phase_portrait}).\\
We found that the fixed points of the nested cycles can be changed and improved when performing full end-to-end training through all solve iterations.\added{ Doing this requires absolutely no new data, the same dataset as from training step 1 can be re-used.} It is possible to set up the cycle-solving programmatically such that a continuous gradient path exists from the beginning to the end, unrolling all intermediate cycle-solving steps \deleted{in between} (fig.~\ref{fig:step_by_step}). We call this method fine-tuning, as it only changes the pre-trained partial models slightly, just enough such that the fixed point moves towards the correct solution. \\
It is known that ML models often have unexpected gradients outside of their trained domain \citep{snyder2013kernels,schmitz2022algorithmic}. Backpropagating through a different type of solver has been attempted before: In Neural Ordinary Differential Equations (Neural ODEs), backpropagation is performed through the ODE solver \citep{chen2018neural}. In Deep Equilibrium Models (DEQ), weight-tied NNs are trained such that from the fixed points of their hidden state the desired target can be calculated easily \citep{bai2019deep}. \\
Our approach is similar: We compute a loss (and perform the according weight update) for the stream after each unit instead of just the flowsheet output, such that the intermediate predictions ideally don't become worse during fine-tuning (note that fig.~\ref{fig:step_by_step} only shows loss computation after the final output stream for better readability). Additionally, we compute a loss after several different amounts of solve iterations. In particular, we first perform 0 solve iterations (using just the initial guesses) and then compute a loss after each unit. Then, in the next forward pass, we perform one solve iteration and again compute the loss after each unit. And so on, for all solve iterations between 0 and 10. Using all solve iterations between 0 and 10 was a design choice and one might as well use less or more. In our experiments, using a range of solve iterations instead of just one had a considerable effect on the result (section \ref{sec:results:finetuning} shows results for three different scenarios).\\
The overall loss for stream $s$ after performing $k$ solve iterations is
\begin{equation}
  \mathcal{L}^{(k,s)} = \mathds{E}_{x^{(s)}} \Big[(\hat{x}^{(s)}_k - x^{(s)})^2 \Big]
\end{equation}
where  $\hat{x}^{(s)}_k$ is the value for the prediction of stream $s$ and solve iteration $k$, and $x^{(s)}$ is the true value for this stream.
Taking into account that in our case each stream is predicted by a series of NNs whose parameters we group into a vector $\theta$, the full loss gradient with which to perform the fine-tuning becomes
\begin{align}
  \frac{\partial \mathcal{L}^{(k,s)}}{\partial \theta} = \
    &\sum_{i=0}^{k}\frac{\partial \mathcal{L}^{(k,s)}}{\partial \hat{x}^{(s)}_i} \
     \frac{\partial \hat{x}^{(s)}_i}{\partial \theta}
\end{align}
The summation over loss iterations in the formula for the loss gradient is because the computation of the value $x^{(s)}_i$ in each iteration $i \in 0 ... k$ contributes to the final value of $\hat{x}^{(s)}_k$ and therefore to the loss and gradient.
After summing the gradients from all amounts of solve iterations, the full gradient update becomes
\begin{equation}
  \Delta \theta = \eta \frac{1}{K} \frac{1}{S} \sum_{k=0}^K \sum_{s=0}^S \frac{\partial \mathcal{L}^{(k,s)}}{\partial \theta}
\end{equation}
where $\eta$ is the learning rate and we normalize by the total number of solve iterations and the number of streams.

\section{Results}
\subsection{Single unit and small cycle}
\begin{figure*}[h]
  \centering
  \includegraphics[width=0.98\linewidth]{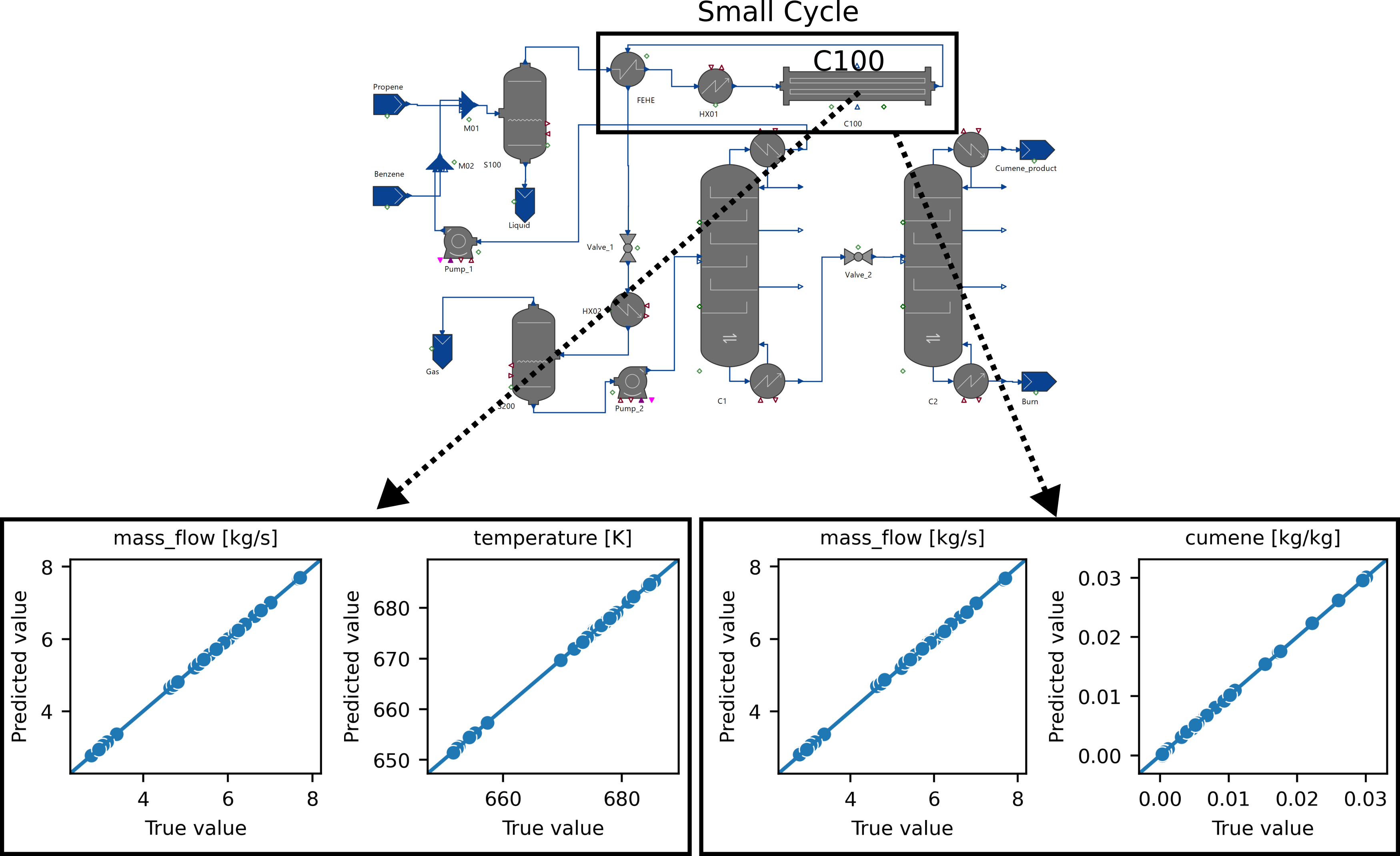}
  \caption{Top: position of the C100 (reactor) and the "small cycle". Bottom left: Parity plot for the single-unit prediction of the C100 reactor. Bottom right: Parity plot for the propagation through the small cycle, including iteratively improving the estimate from initial values. The second variable shown is Cumene and not Temperature, because the temperature is constant at the outlet of the FEHE heat exchanger.}
  \label{fig:single_unit_and_small_cycle_parities}
\end{figure*}
\label{sec:single_unit_and_small_cycle_results}
In order to show that failures during the cycle solving were not due to the single unit models, we first present results for the single unit prediction and for a small cycle within the Cumene process (fig.~\ref{fig:cumene_process}). \\
The most complex process unit in the flowsheet is the reactor C100. Even for the C100, the predictions are almost perfect (fig.~\ref{fig:single_unit_and_small_cycle_parities}). Notice that the near-perfect accuracy was only achievable because we are using simulated data as a substitute for real-world plant sensor data. The predictions for all other units were similarly good (see table \ref{table:single_unit_errors}). Based on this near-perfect accuracy on the single-unit prediction, one would expect that combining all single-unit ML models into a full "flowsheet" would yield similarly good results on the end-to-end prediction task.\\
\deleted{Since all training data are from the steady-state operation of the plant (see section \ref{sec:data_generation}), one can only expect good prediction accuracy in steady-state conditions. During the cycle solving, however, the model may have to make predictions outside of this training data domain. In particular, when injecting an initial guess for the tear stream (as an example, stream $s_{3,1}$ in fig.~\ref{fig:tear_stream}), this initial guess is not a solution of the stationary state in combination with the flowsheet inputs (hence that the cycle solving is necessary in the first place). Therefore, this combination of flowsheet input and cyce-stream-initial-guess has not been seen during single-unit training. \\}

As a first experiment with cycle solving with ML-based models, we tried to solve the small cycle in fig.~\ref{fig:single_unit_and_small_cycle_parities} (for a larger version of the flowsheet see fig.~\ref{fig:cumene_process}). The cycle was extracted from the Cumene process flowsheet and treated as an independent flowsheet. \\
The three units \textit{FEHE}, \textit{HX01} and \textit{C100} were trained independently (single-unit training, training step 1 in fig.~\ref{fig:step_by_step}). After training each of the three units, the models were connected to obtain one overall model representing the entire flowsheet. The input to this new flowsheet is the non-cycle input to the \textit{FEHE} heat exchanger unit (the \textit{FEHE} has two inputs, the other being the tear stream coming from \textit{C100}). We tried cycle-solving methods as introduced in sec. \ref{sec:cycle_solving_methods} to solve this cycle. All four methods behaved almost exactly the same and converged towards the same (correct) value (fig.~\ref{fig:single_unit_and_small_cycle_parities}), therefore per-method results are not shown here. This result indicates that the tested cycle-solving methods are generally suitable for ML model-based cycles, a finding which replicates results from previous studies (e.g.~\citet{bubel2021modular}, who used the Newton solver). \\

\subsection{Full flowsheet before fine-tuning}
\label{sec:results:before_finetuning}
 We tried the same four cycle-solving methods that worked well for the small cycle in section~\ref{sec:single_unit_and_small_cycle_results} on the entire flowsheet, which contains a large cycle and a smaller cycle nested within this larger cycle.
The scalar error we report for the full flowsheet is measured as the mean standard deviation-scaled difference between the true values and the prediction at the top output of the C2 column. This output is the one that produces the Cumene and therefore is the most relevant.\\
\begin{figure}[h]
  \centering
  \includegraphics[width=0.99\linewidth]{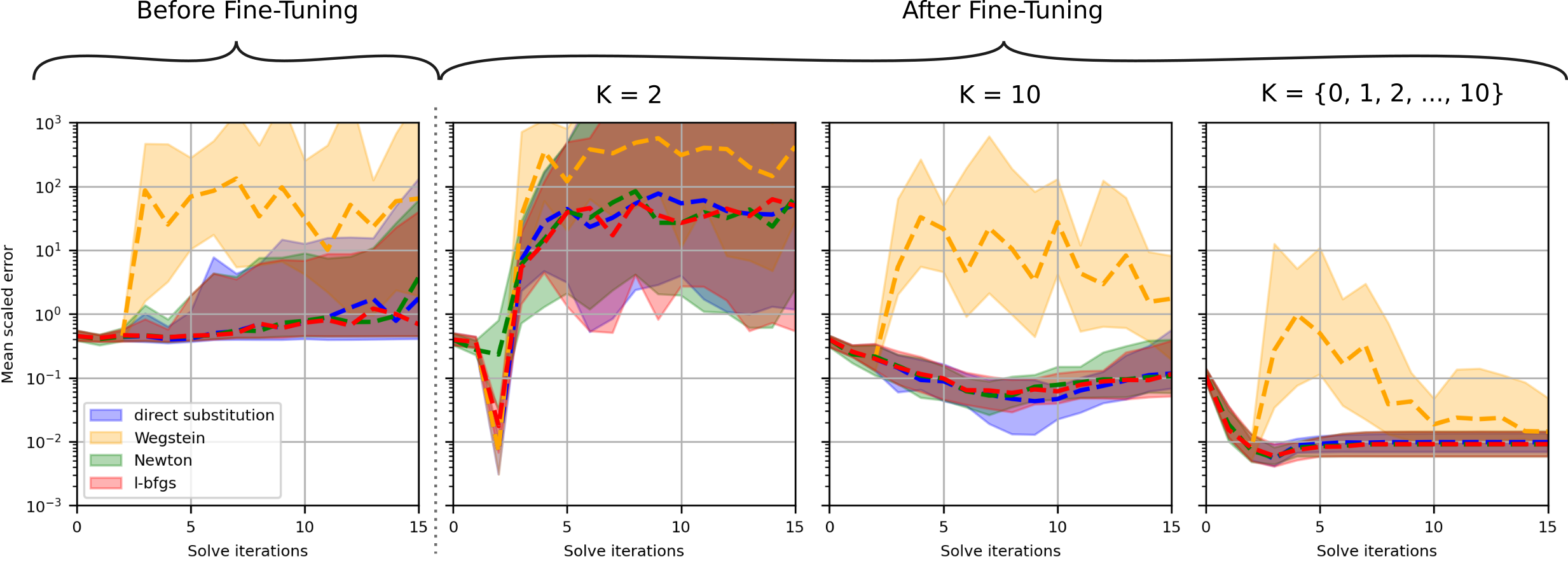}
  \caption{Solving the full flowsheet (with both cycles). The y-axis represents the mean, standard deviation-scaled error of the prediction of the C2-top output. The dashed lines represent the 50th percentile. The colored ranges represent the 25th to 75th percentile. Left: Without fine-tuning. Center left: fine-tuning only for $K=2$ solve iterations. Center right: fine-tuning for $K=10$ solve iterations. Right: fine-tuning for $K=\{0,1,2,...,10\}$ solve iterations. Note the difference between fine-tuning solve iterations (indicated by different plots) and model initialization (prediction) solve iterations (indicated by different colors within each plot).}
  \label{fig:convergence_combined}
\end{figure}

Unlike the results from solving just the small cycle, where established cycle-solving techniques worked well, the same techniques did not work for the full flowsheet (fig.~\ref{fig:convergence_combined} left plot and table~\ref{table:single_unit_errors} bottom row). With an increasing number of solve iterations, the predictions became worse, in other words, the solving procedure diverged. The shaded areas within the plots often overlap, making them indistinguishable from one another; however this does not matter as the main result displayed in the plot is that none of the solve methods perform well before fine-tuning. The Wegstein method is usually regarded as a particularly stable method, but in our experiments the opposite was true.  The Wegstein method typically performed worse than the other methods, and its variability between experiments was high.\\
\deleted{A qualitative analysis of the solving behavior revealed that divergence happens especially in regions of the input domain where lesser data was available (section \ref{sec:results:phase_portrait}). What was surprising was that all four examined cycle-solving methods performed similarly.}

\subsection{Full flowsheet after fine-tuning}
\label{sec:results:finetuning}
\begin{figure}[h]
  \centering
  \includegraphics[width=0.6\linewidth]{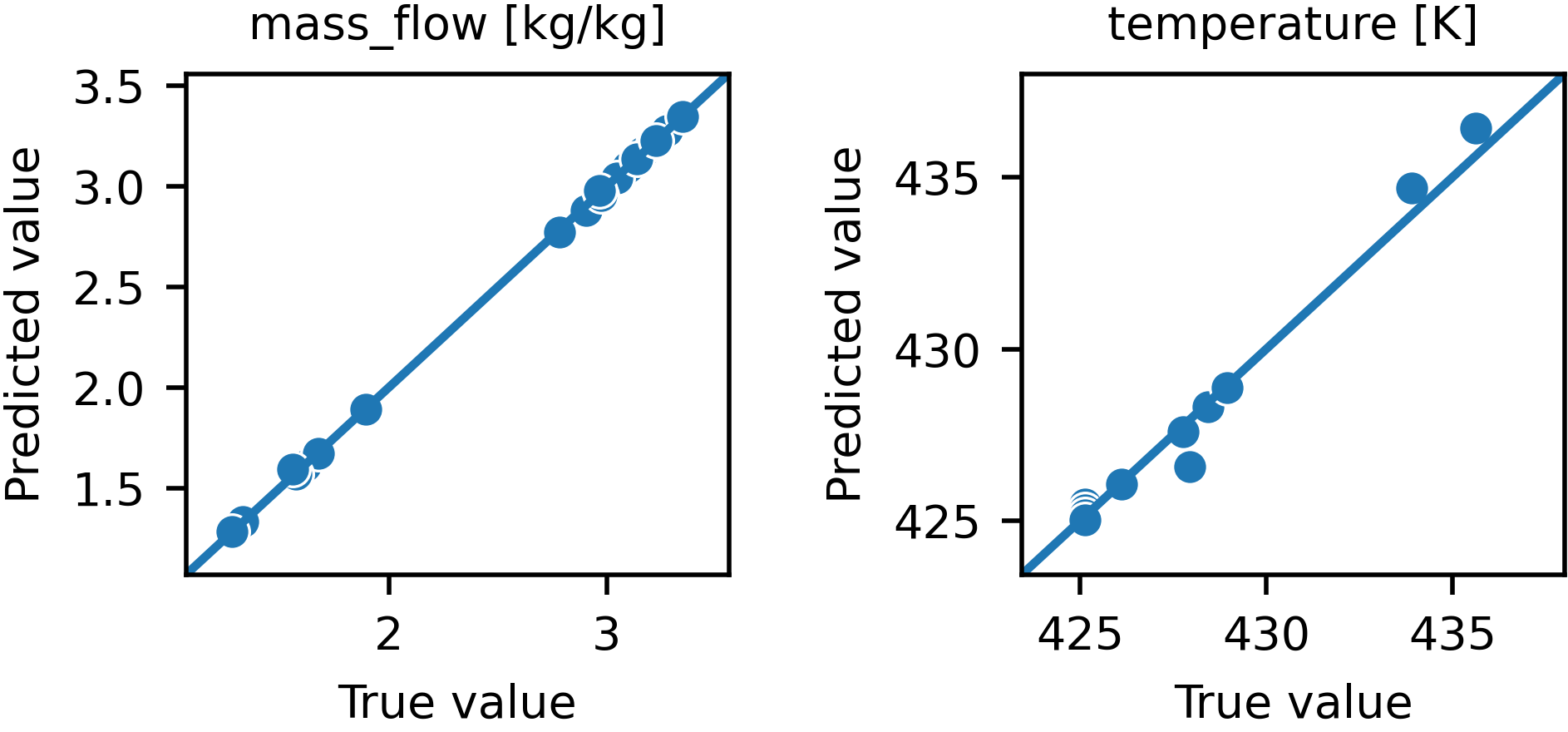}
  \caption{Parity plot of the end-to-end prediction of the entire flowsheet after fine-tuning.}
  \label{fig:parity_finetuning}
\end{figure}

The instabilities of the cycle-solving methods can be avoided by using the fine-tuning method introduced in this study (Training step 2, see section \ref{sec:methods:finetuning}).\added{ Note that absolutely no new training data has been used to do this. Fine-tuning can re-use the existing training data used for training step 1.} The effect of applying fine-tuning on the accuracy of the model is visualized in fig.~\ref{fig:convergence_combined}. In order to interpret the plots in fig.~\ref{fig:convergence_combined} correctly, it is important to distinguish between solve iterations during \textit{training} (for fine-tuning, as shown in fig.~\ref{fig:step_by_step} bottom), and during \textit{evaluation} of the test data. In both cases, the cycles are solved by iterating over the state variables. The difference is that during training, this solving procedure is done on the training set and for the purpose of optimizing the NN parameters, whereas during evaluation, it is done on the (previously unseen) test set in order to evaluate the model's performance. \\
The fine-tuning can be done for a fixed number of total solve iterations $K$, or for several different values of $K$. For instance, we can do fine-tuning with $K=2$. What that means is that during fine-tuning, we alternate between doing a forward pass through the flowsheet, including going through the cycle 2 times, and doing a backward pass as indicated in fig.~\ref{fig:step_by_step}. \\
For the result of performing fine-tuning with $K=2$ and $K=10$, see fig.~\ref{fig:convergence_combined} center left and center right. Note that the solve iterations on the x-axis corrspond to the \textit{prediction} solve iterations. We can also set $K$ to a list of values, for instance $K=\{0, 1, 2, 3, ..., 10\}$. In this case, we repeatedly iterate between each value of $K$ and perform one fine-tuning iteration with each $K$. We found that doing fine-tuning for several total solve iterations iteratively worked best (fig.~\ref{fig:convergence_combined} right). \\
Interestingly, if fine-tuning is only done for a specific number of solve iterations, the prediction works best for that number of solve iterations too. For instance, in the case $K=2$ during fine-tuning, the \textit{prediction} with $2$ solve iterations worked best (fig.~\ref{fig:convergence_combined} center left). This was not the case anymore when doing fine-tuning with 10 solve iterations, however (fig.~\ref{fig:convergence_combined} center right). After fine-tuning in the $K={0, 1, 2, ... 10}$ condition, predictions were near-perfect (fig.~\ref{fig:parity_finetuning}). Fine-tuning for the results in fig.~\ref{fig:convergence_combined} was always done with the direct substitution method, regardless of the prediction solve method, because it worked the best, but see the appendix \ref{sec:appendix:fine_tuning} for all combinations of solve methods for training and prediction.

\subsubsection{Phase portrait of cycle solving}
\label{sec:results:phase_portrait}
\begin{figure*}[h]
  \centering
  \includegraphics[width=0.99\linewidth]{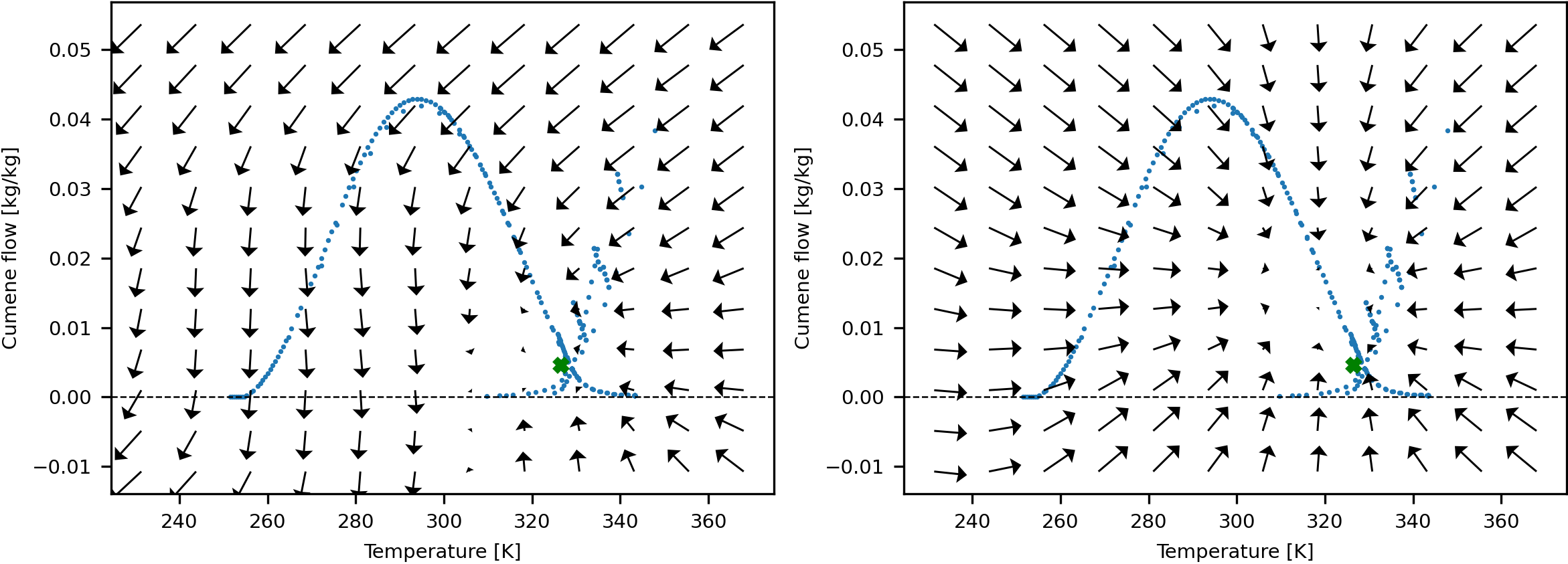}
  \caption{Phase portrait of temperature [K] and Cumene content [kg/kg]. Arrows indicate the direction of the negative gradient of the flowsheet response function (eq.~\ref{eq:flowsheet_response}). The green star indicates the true value for the cycle. The blue dots indicate the data points. Left: Before fine-tuning. Right: After fine-tuning.}
  \label{fig:phase_portrait}
\end{figure*}

To further illustrate the difference in the flowsheet response before and after fine-tuning, we display a phase portrait in fig.~\ref{fig:phase_portrait}. The data distribution looks unusual compared to familiar ML datasets due to the fact that in the simulation for the data generation, only a narrow range of values is permissible.Before fine-tuning, the gradient points in seemingly random directions, which explains why the cycle-solving methods which indirectly or directly depend on the gradient do not converge towards the true solution.\\
After fine-tuning, it can be seen that  from all locations in the diagram, the gradient points towards the correct solution. This indicates that in order for the solver to converge towards the true stationary state (the one corresponding to the steady-state of the plant), it is not enough to have accurate single-unit models. As seen in table~\ref{table:single_unit_errors}, the accuracy of the single-unit models before fine-tuning is excellent. Despite this, their gradient along the nested cycles is not useful for the solver. In addition to good accuracy, it is also necessary to constrain the gradient.

\subsubsection{Fine-tuning impact on single-unit accuracy}
An interesting question is whether the single-unit accuracy gets impeded by the fine-tuning. This would be a trade-off between single-unit accuaracy and performance of the end-to-end prediction. \\
Table \ref{table:single_unit_errors} shows the single-unit accuracy before and after fine-tuning. While most units had an almost negligible decrease in accuracy, the unit S100 is an exception and its $r^2$-value decreased from $0.999$ to $0.585$. We do not know why the unit S100 changed its prediction by so much while all other units are virtually unchanged. To further investigate, we performed another experiment. For training step 1, we proceeded as usual, but in training step 2, the fine-tuning was done with the weights of the S100 unit frozen. Table \ref{table:single_unit_errors} shows the results of this experiment. As expected, the performance of the S100 unit is unchanged ($r^2=0.999$ before and after fine-tuning), but interestingly, there is no one single unit that "takes over" the role of the S100 and significantly decreases in accuracy. The full-flowsheet predictions are still near-perfect even with the frozen S100 model included (results not shown, but almost indistinguishable from results in fig.~\ref{fig:parity_finetuning}). This indicates that the relatively large change in the S100 accuracy before was not necessary to induce stable and correct fixed points.\\
\begin{table*}
  \begin{tabular}{llrrr}
    Unit Name  & Unit Type           & \pbox{20cm}{After only \\ Training step 1} & After fine-tuning & \pbox{20cm}{After fine-tuning \\ with frozen S100}  \\
    \hline
    C1         & Distillation column & 1.000                & 0.997             & 0.996                         \\
    C100       & Tubular reactor     & 0.999                & 0.997             & 0.997                         \\
    C2         & Distillation column & 1.000                & 0.996             & 0.994                         \\
    FEHE       & Heat exchanger      & 1.000                & 0.984             & 0.970                         \\
    HX01       & Heater              & 1.000                & 0.997             & 0.996                         \\
    HX02       & Cooler              & 1.000                & 0.994             & 0.993                         \\
    M01        & Mixer               & 1.000                & 0.956             & \textbf{0.942}                \\
    M02        & Mixer               & 0.999                & 0.966             & 0.968                         \\
    Pump1      & Pump                & 0.999                & 0.995             & 0.994                         \\
    Pump2      & Pump                & 1.000                & 1.000             & 1.000                         \\
    S100       & Separator           & 0.999                & \textbf{0.585}    & 0.999                         \\
    S200       & Separator           & 1.000                & 0.999             & 0.999                         \\
    Valve1     & Valve               & \textbf{0.998}       & 0.975             & 0.978                         \\
    Valve2     & Valve               & 1.000                & 1.000             & 0.999                         \\
    \hline
    \pbox{20cm}{End-to-end \\ prediction} & Whole Flowsheet     & -0.806               & 0.993             & 0.988                         \\
  \end{tabular}
  \caption{The performance impact of fine-tuning on the single-unit and end-to-end prediction performance. The prediction accuracy is measured with the coefficient of determination ($r^2$) before and after fine-tuning. Bold text indicates the lowest value per column (excluding end-to-end prediction). Note that the performance of the end-to-end prediction before fine-tuning varies (see fig.\ref{fig:convergence_combined}).}
  \label{table:single_unit_errors}
\end{table*}

\section{Discussion}
The current study is the result of trying to apply published methods of ML-based chemical plant models to an industrial-scale example, in our case the Cumene process. It was found that published methods work well on smaller scales, but do not easily scale up. In particular, we noticed that in the presence of nested cycles, which are common in chemical process engineering, the cycle-solving operation runs into instabilities. All four cycle-solving methods we tested suffered from these instabilities. We showed how the gradient of the flowsheet response function (eq.~\ref{eq:flowsheet_response}) is not suitable at least for gradient-based solve methods. \\
To mitigate this problem, we introduce a method we call fine-tuning to condition the pre-trained networks such that cycle solving with classical methods works again. By backpropagating the training loss through the iterative solving procedure, we were able to tune the NNs ever so slightly such that the instabilities in the solving operation are overcome. We showed that after fine-tuning, the gradient of the flowsheet response function "points" directly towards the correct stationary solution and hence makes solving the cycles trivial.  \\
Looking ahead, the next obvious step is to apply our method to a real plant and to learn from sensor data. To do this, there is one hurdle: there won't be sensor data available for all intermediate values. Therefore, training the ML models as we did so far would not be possible. This problem can be overcome with a hybrid approach in which one trains from both simulated and sensor data. When training with sensor data, error gradients can be generated from the variables available, and the other variables can be ignored. The simulated data are then used to train those variables for which no sensor data are available. \\
We also see a possible second application of the proposed fine-tuning method here: by letting the error gradient flow end-to-end, intermediate units for which data are not available can be learned implicitly. This only works if data for at most every second unit is missing. Alternatively, if indeed data for more than every second unit is missing, one can group sequential missing units into one meta-model, and then apply the fine-tuning method as before.\added{ Introducing meta-models makes the fine-grained control over the simulation that a structured model offers less precise, but missing data can necessitate it. One can think of the number and size of meta-models as an adjustable compromise between the fully fine-grained control over the entire process and the reduced-variable data needs of a black-box end-to-end model.} \\
\deleted{The fine-tuning is also interesting on its own: in a real-world scenario, one could measure from an existing plant and get the "real world steady state". If one has an already fitted model, it is unlikely that its stationary solution is accurate with regards to the new data, because the single-unit models are not representing the degraded state of the system. The fine-tuning method is able to easily tune the models to reflect the actually measured steady-state.\\}
Another promising direction for future work is to use fine-tuned ML models for optimization. Mathematically, ML-based models have properties such as smoothness, no discontinuities, and availability of gradients which make them well-suited for optimization. Also, ML models are typically much faster to evaluate than solving first-principles-based equation systems.\deleted{ While we did not explore optimization with ML-based models in the current study, there is enormous research interest in this direction \mbox{\citep{schweidtmann2018machine,bradford2018efficient,asprion2019gray,schweidtmann2019deterministic,burre2022global}} and} Our proposed fine-tuning method might pave the way to optimization of larger and more complex plants.\\
In a nutshell, we show here that when extending structured ML-based plant simulation to industrial scales, existing methods run into instabilities, and one way to solve this problem is by fine-tuning the partial models end-to-end.

\section*{Acknowledgements}
ME and MG work at the BASLEARN-TU Berlin/BASF Joint Lab for Machine Learning, co-financed by TU Berlin and BASF SE.
We thank Bruno Betoni Parodi for fruitful discussions and for managing BASLEARN.

\bibliographystyle{elsarticle-harv}
\bibliography{references}

\begin{thebibliography}{32}
\expandafter\ifx\csname natexlab\endcsname\relax\def\natexlab#1{#1}\fi
\providecommand{\url}[1]{\texttt{#1}}
\providecommand{\href}[2]{#2}
\providecommand{\path}[1]{#1}
\providecommand{\DOIprefix}{doi:}
\providecommand{\ArXivprefix}{arXiv:}
\providecommand{\URLprefix}{URL: }
\providecommand{\Pubmedprefix}{pmid:}
\providecommand{\doi}[1]{\href{http://dx.doi.org/#1}{\path{#1}}}
\providecommand{\Pubmed}[1]{\href{pmid:#1}{\path{#1}}}
\providecommand{\bibinfo}[2]{#2}
\ifx\xfnm\relax \def\xfnm[#1]{\unskip,\space#1}\fi
\bibitem[{Bai et~al.(2019)Bai, Kolter and Koltun}]{bai2019deep}
\bibinfo{author}{Bai, S.}, \bibinfo{author}{Kolter, J.Z.},
  \bibinfo{author}{Koltun, V.}, \bibinfo{year}{2019}.
\newblock \bibinfo{title}{Deep equilibrium models}.
\newblock \bibinfo{journal}{Advances in Neural Information Processing Systems}
  \bibinfo{volume}{32}.
\bibitem[{Banach(1922)}]{banach1922operations}
\bibinfo{author}{Banach, S.}, \bibinfo{year}{1922}.
\newblock \bibinfo{title}{Sur les op{\'e}rations dans les ensembles abstraits
  et leur application aux {\'e}quations int{\'e}grales}.
\newblock \bibinfo{journal}{Fundamenta mathematicae} \bibinfo{volume}{3},
  \bibinfo{pages}{133--181}.
\bibitem[{Bikmukhametov and J{\"a}schke(2020)}]{bikmukhametov2020combining}
\bibinfo{author}{Bikmukhametov, T.}, \bibinfo{author}{J{\"a}schke, J.},
  \bibinfo{year}{2020}.
\newblock \bibinfo{title}{Combining machine learning and process engineering
  physics towards enhanced accuracy and explainability of data-driven models}.
\newblock \bibinfo{journal}{Computers \& Chemical Engineering}
  \bibinfo{volume}{138}, \bibinfo{pages}{106834}.
\bibitem[{Bogojeski et~al.(2021)Bogojeski, Sauer, Horn and
  M{\"u}ller}]{bogojeski2021forecasting}
\bibinfo{author}{Bogojeski, M.}, \bibinfo{author}{Sauer, S.},
  \bibinfo{author}{Horn, F.}, \bibinfo{author}{M{\"u}ller, K.R.},
  \bibinfo{year}{2021}.
\newblock \bibinfo{title}{Forecasting industrial aging processes with machine
  learning methods}.
\newblock \bibinfo{journal}{Computers \& Chemical Engineering}
  \bibinfo{volume}{144}, \bibinfo{pages}{107123}.
\bibitem[{Bongartz and Mitsos(2019)}]{bongartz2019deterministic}
\bibinfo{author}{Bongartz, D.}, \bibinfo{author}{Mitsos, A.},
  \bibinfo{year}{2019}.
\newblock \bibinfo{title}{Deterministic global flowsheet optimization: Between
  equation-oriented and sequential-modular methods}.
\newblock \bibinfo{journal}{AIChE Journal} \bibinfo{volume}{65},
  \bibinfo{pages}{1022--1034}.
\bibitem[{Briceno-Mena et~al.(2023)Briceno-Mena, Arges and
  Romagnoli}]{briceno2023machine}
\bibinfo{author}{Briceno-Mena, L.A.}, \bibinfo{author}{Arges, C.G.},
  \bibinfo{author}{Romagnoli, J.A.}, \bibinfo{year}{2023}.
\newblock \bibinfo{title}{Machine learning-based surrogate models and transfer
  learning for derivative free optimization of ht-pem fuel cells}.
\newblock \bibinfo{journal}{Computers \& Chemical Engineering}
  \bibinfo{volume}{171}, \bibinfo{pages}{108159}.
\bibitem[{Bubel et~al.(2021)Bubel, Ludl, Seidel, Asprion and
  Bortz}]{bubel2021modular}
\bibinfo{author}{Bubel, M.}, \bibinfo{author}{Ludl, P.O.},
  \bibinfo{author}{Seidel, T.}, \bibinfo{author}{Asprion, N.},
  \bibinfo{author}{Bortz, M.}, \bibinfo{year}{2021}.
\newblock \bibinfo{title}{A modular approach for surrogate modeling of
  flowsheets}.
\newblock \bibinfo{journal}{Chemie Ingenieur Technik} \bibinfo{volume}{93},
  \bibinfo{pages}{1987--1997}.
\bibitem[{Burre et~al.(2022)Burre, Kabatnik, Al-Khatib, Bongartz, Jupke and
  Mitsos}]{burre2022global}
\bibinfo{author}{Burre, J.}, \bibinfo{author}{Kabatnik, C.},
  \bibinfo{author}{Al-Khatib, M.}, \bibinfo{author}{Bongartz, D.},
  \bibinfo{author}{Jupke, A.}, \bibinfo{author}{Mitsos, A.},
  \bibinfo{year}{2022}.
\newblock \bibinfo{title}{Global flowsheet optimization for reductive
  dimethoxymethane production using data-driven thermodynamic models}.
\newblock \bibinfo{journal}{Computers \& Chemical Engineering}
  \bibinfo{volume}{162}, \bibinfo{pages}{107806}.
\bibitem[{Byrne and Bogle(2000)}]{byrne2000global}
\bibinfo{author}{Byrne, R.}, \bibinfo{author}{Bogle, I.}, \bibinfo{year}{2000}.
\newblock \bibinfo{title}{Global optimization of modular process flowsheets}.
\newblock \bibinfo{journal}{Industrial \& engineering chemistry research}
  \bibinfo{volume}{39}, \bibinfo{pages}{4296--4301}.
\bibitem[{Caballero and Grossmann(2008)}]{caballero2008algorithm}
\bibinfo{author}{Caballero, J.A.}, \bibinfo{author}{Grossmann, I.E.},
  \bibinfo{year}{2008}.
\newblock \bibinfo{title}{An algorithm for the use of surrogate models in
  modular flowsheet optimization}.
\newblock \bibinfo{journal}{AIChE journal} \bibinfo{volume}{54},
  \bibinfo{pages}{2633--2650}.
\bibitem[{Casella and Bachmann(2021)}]{casella2021choice}
\bibinfo{author}{Casella, F.}, \bibinfo{author}{Bachmann, B.},
  \bibinfo{year}{2021}.
\newblock \bibinfo{title}{On the choice of initial guesses for the
  newton-raphson algorithm}.
\newblock \bibinfo{journal}{Applied Mathematics and Computation}
  \bibinfo{volume}{398}, \bibinfo{pages}{125991}.
\bibitem[{Chen et~al.(2018)Chen, Rubanova, Bettencourt and
  Duvenaud}]{chen2018neural}
\bibinfo{author}{Chen, R.T.}, \bibinfo{author}{Rubanova, Y.},
  \bibinfo{author}{Bettencourt, J.}, \bibinfo{author}{Duvenaud, D.K.},
  \bibinfo{year}{2018}.
\newblock \bibinfo{title}{Neural ordinary differential equations}.
\newblock \bibinfo{journal}{Advances in neural information processing systems}
  \bibinfo{volume}{31}.
\bibitem[{Crowe and Nishio(1975)}]{crowe1975convergence}
\bibinfo{author}{Crowe, C.M.}, \bibinfo{author}{Nishio, M.},
  \bibinfo{year}{1975}.
\newblock \bibinfo{title}{Convergence promotion in the simulation of chemical
  processes—the general dominant eigenvalue method}.
\newblock \bibinfo{journal}{AIChE Journal} \bibinfo{volume}{21},
  \bibinfo{pages}{528--533}.
\bibitem[{Fletcher(2013)}]{fletcher2013practical}
\bibinfo{author}{Fletcher, R.}, \bibinfo{year}{2013}.
\newblock \bibinfo{title}{Practical methods of optimization}.
\newblock \bibinfo{publisher}{John Wiley \& Sons}.
\bibitem[{Gordon and Pistikopoulos(2022)}]{gordon2022data}
\bibinfo{author}{Gordon, C.A.K.}, \bibinfo{author}{Pistikopoulos, E.N.},
  \bibinfo{year}{2022}.
\newblock \bibinfo{title}{Data-driven and safety-aware holistic production
  planning}.
\newblock \bibinfo{journal}{Journal of Loss Prevention in the Process
  Industries} \bibinfo{volume}{77}, \bibinfo{pages}{104754}.
\bibitem[{Heese et~al.(2019)Heese, Walczak, Seidel, Asprion and
  Bortz}]{heese2019optimized}
\bibinfo{author}{Heese, R.}, \bibinfo{author}{Walczak, M.},
  \bibinfo{author}{Seidel, T.}, \bibinfo{author}{Asprion, N.},
  \bibinfo{author}{Bortz, M.}, \bibinfo{year}{2019}.
\newblock \bibinfo{title}{Optimized data exploration applied to the simulation
  of a chemical process}.
\newblock \bibinfo{journal}{Computers \& Chemical Engineering}
  \bibinfo{volume}{124}, \bibinfo{pages}{326--342}.
\bibitem[{Kingma and Ba(2014)}]{kingma2014adam}
\bibinfo{author}{Kingma, D.P.}, \bibinfo{author}{Ba, J.}, \bibinfo{year}{2014}.
\newblock \bibinfo{title}{Adam: A method for stochastic optimization}.
\newblock \bibinfo{journal}{arXiv preprint arXiv:1412.6980} .
\bibitem[{Lee et~al.(2018)Lee, Shin and Realff}]{lee2018machine}
\bibinfo{author}{Lee, J.H.}, \bibinfo{author}{Shin, J.},
  \bibinfo{author}{Realff, M.J.}, \bibinfo{year}{2018}.
\newblock \bibinfo{title}{Machine learning: Overview of the recent progresses
  and implications for the process systems engineering field}.
\newblock \bibinfo{journal}{Computers \& Chemical Engineering}
  \bibinfo{volume}{114}, \bibinfo{pages}{111--121}.
\bibitem[{Ludl et~al.(2022)Ludl, Heese, H{\"o}ller, Asprion and
  Bortz}]{ludl2022using}
\bibinfo{author}{Ludl, P.O.}, \bibinfo{author}{Heese, R.},
  \bibinfo{author}{H{\"o}ller, J.}, \bibinfo{author}{Asprion, N.},
  \bibinfo{author}{Bortz, M.}, \bibinfo{year}{2022}.
\newblock \bibinfo{title}{Using machine learning models to explore the solution
  space of large nonlinear systems underlying flowsheet simulations with
  constraints}.
\newblock \bibinfo{journal}{Frontiers of Chemical Science and Engineering}
  \bibinfo{volume}{16}, \bibinfo{pages}{183--197}.
\bibitem[{Luyben(2010)}]{luyben2010design}
\bibinfo{author}{Luyben, W.L.}, \bibinfo{year}{2010}.
\newblock \bibinfo{title}{Design and control of the cumene process}.
\newblock \bibinfo{journal}{Industrial \& engineering chemistry research}
  \bibinfo{volume}{49}, \bibinfo{pages}{719--734}.
\bibitem[{Orbach and Crowe(1971)}]{orbach1971convergence}
\bibinfo{author}{Orbach, O.}, \bibinfo{author}{Crowe, C.},
  \bibinfo{year}{1971}.
\newblock \bibinfo{title}{Convergence promotion in the simulation of chemical
  processes with recycle-the dominant eigenvalue method}.
\newblock \bibinfo{journal}{The Canadian Journal of Chemical Engineering}
  \bibinfo{volume}{49}, \bibinfo{pages}{509--513}.
\bibitem[{Palmer and Realff(2002a)}]{palmer2002metamodeling}
\bibinfo{author}{Palmer, K.}, \bibinfo{author}{Realff, M.},
  \bibinfo{year}{2002}a.
\newblock \bibinfo{title}{Metamodeling approach to optimization of steady-state
  flowsheet simulations: Model generation}.
\newblock \bibinfo{journal}{Chemical Engineering Research and Design}
  \bibinfo{volume}{80}, \bibinfo{pages}{760--772}.
\bibitem[{Palmer and Realff(2002b)}]{palmer2002optimization}
\bibinfo{author}{Palmer, K.}, \bibinfo{author}{Realff, M.},
  \bibinfo{year}{2002}b.
\newblock \bibinfo{title}{Optimization and validation of steady-state flowsheet
  simulation metamodels}.
\newblock \bibinfo{journal}{Chemical Engineering Research and Design}
  \bibinfo{volume}{80}, \bibinfo{pages}{773--782}.
\bibitem[{Pho and Lapidus(1973)}]{pho1973tear}
\bibinfo{author}{Pho, T.}, \bibinfo{author}{Lapidus, L.}, \bibinfo{year}{1973}.
\newblock \bibinfo{title}{Topics in computer-aided design: Part i. an optimum
  tearing algorithm for recycle systems}.
\newblock \bibinfo{journal}{AIChE Journal} \bibinfo{volume}{19},
  \bibinfo{pages}{1170--1181}.
\bibitem[{Schmitz et~al.(2022)Schmitz, M{\"u}ller and
  Chmiela}]{schmitz2022algorithmic}
\bibinfo{author}{Schmitz, N.F.}, \bibinfo{author}{M{\"u}ller, K.R.},
  \bibinfo{author}{Chmiela, S.}, \bibinfo{year}{2022}.
\newblock \bibinfo{title}{Algorithmic differentiation for automated modeling of
  machine learned force fields}.
\newblock \bibinfo{journal}{The Journal of Physical Chemistry Letters}
  \bibinfo{volume}{13}, \bibinfo{pages}{10183--10189}.
\bibitem[{Schweidtmann et~al.(2022)Schweidtmann, Bongartz and
  Mitsos}]{schweidtmann2022optimization}
\bibinfo{author}{Schweidtmann, A.M.}, \bibinfo{author}{Bongartz, D.},
  \bibinfo{author}{Mitsos, A.}, \bibinfo{year}{2022}.
\newblock \bibinfo{title}{Optimization with trained machine learning models
  embedded}, in: \bibinfo{booktitle}{Encyclopedia of Optimization}.
  \bibinfo{publisher}{Springer}, pp. \bibinfo{pages}{1--8}.
\bibitem[{Shacham et~al.(1982)Shacham, Macchieto, Stutzman and
  Babcock}]{shacham1982equation}
\bibinfo{author}{Shacham, M.}, \bibinfo{author}{Macchieto, S.},
  \bibinfo{author}{Stutzman, L.}, \bibinfo{author}{Babcock, P.},
  \bibinfo{year}{1982}.
\newblock \bibinfo{title}{Equation oriented approach to process flowsheeting}.
\newblock \bibinfo{journal}{Computers \& Chemical Engineering}
  \bibinfo{volume}{6}, \bibinfo{pages}{79--95}.
\bibitem[{Shalaby et~al.(2021)Shalaby, Elkamel, Douglas, Zhu and
  Zheng}]{shalaby2021machine}
\bibinfo{author}{Shalaby, A.}, \bibinfo{author}{Elkamel, A.},
  \bibinfo{author}{Douglas, P.L.}, \bibinfo{author}{Zhu, Q.},
  \bibinfo{author}{Zheng, Q.P.}, \bibinfo{year}{2021}.
\newblock \bibinfo{title}{A machine learning approach for modeling and
  optimization of a co2 post-combustion capture unit}.
\newblock \bibinfo{journal}{Energy} \bibinfo{volume}{215},
  \bibinfo{pages}{119113}.
\bibitem[{Smith(2005)}]{smith2005chemical}
\bibinfo{author}{Smith, R.}, \bibinfo{year}{2005}.
\newblock \bibinfo{title}{Chemical process: design and integration}.
\newblock \bibinfo{publisher}{John Wiley \& Sons}.
\bibitem[{Snyder et~al.(2013)Snyder, Mika, Burke and
  M{\"u}ller}]{snyder2013kernels}
\bibinfo{author}{Snyder, J.C.}, \bibinfo{author}{Mika, S.},
  \bibinfo{author}{Burke, K.}, \bibinfo{author}{M{\"u}ller, K.R.},
  \bibinfo{year}{2013}.
\newblock \bibinfo{title}{Kernels, pre-images and optimization}.
\newblock \bibinfo{journal}{Empirical Inference: Festschrift in Honor of
  Vladimir N. Vapnik} , \bibinfo{pages}{245--259}.
\bibitem[{Wegstein(1958)}]{wegstein1958accelerating}
\bibinfo{author}{Wegstein, J.H.}, \bibinfo{year}{1958}.
\newblock \bibinfo{title}{Accelerating convergence of iterative processes}.
\newblock \bibinfo{journal}{Communications of the ACM} \bibinfo{volume}{1},
  \bibinfo{pages}{9--13}.
\bibitem[{Zapf and Wallek(2021)}]{zapf2021gray}
\bibinfo{author}{Zapf, F.}, \bibinfo{author}{Wallek, T.}, \bibinfo{year}{2021}.
\newblock \bibinfo{title}{Gray-box surrogate models for flash, distillation and
  compression units of chemical processes}.
\newblock \bibinfo{journal}{Computers \& Chemical Engineering}
  \bibinfo{volume}{155}, \bibinfo{pages}{107510}.

\end{thebibliography}

\appendix
\section{Training details}
\label{sec:training_details}
Each NN had two hidden layers with 100 neurons each, and a skip connection layer going directly from the input to the output. The activation function was the softplus in all cases:
\begin{equation}
  a(x) = \log(1+e^x)
\end{equation}
Single-unit training was done with the Adam optimizer \citep{kingma2014adam} with a learning rate of $2.0e-5$ for $2e4$ epochs.
Fine-tuning training was also done with the Adam optimizer, for $500$ epochs with a learning rate of $2.0e-5$.
Input to each NN was normalized with the mean $\mu$ and the standard deviation $\sigma$:
\begin{equation}
  f_{normal}(x) = \frac{x - \mu}{\sigma}
\end{equation}
After prediction, the reverse of the normalizing operation was performed
\begin{equation}
  f_{reverse}(x) = x * \sigma + \mu
\end{equation}

\section{Further fine-tuning results}
\label{sec:appendix:fine_tuning}
\begin{figure}[h]
  \centering
  \includegraphics[width=0.99\linewidth]{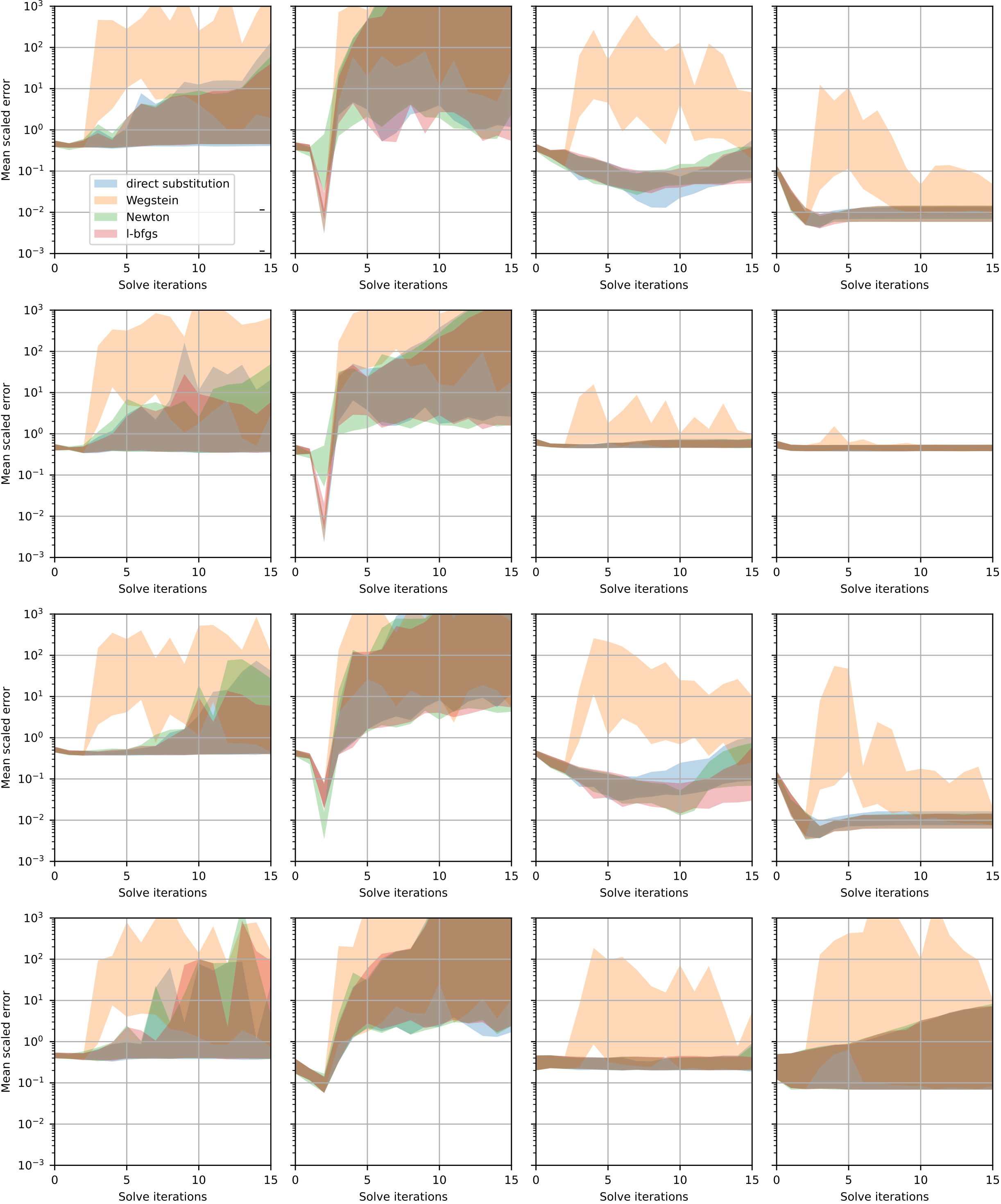}
  \caption{Each row of plots corresponds to fine-tuning with one particular solve method, from top to bottom \textit{direct substitution}, \textit{Wegstein}, \textit{Newton}, \textit{L-BFGS}. Each column of plots corresponds to fine-tuning with the following fine-tuning conditions (left to right): \textit{no fine-tuning}, \textit{fine-tuning 2 solve iterations}, \textit{fine-tuning 10 solve iterations}, \textit{fine-tuning $0, 1, ..., 10$ solve iterations}. The colors within each plot represent the solve method used for prediction (see the top left plot for a legend).}
  \label{fig:convergence_extended}
\end{figure}

It is possible to use a different solve method for training than for inference. In the main text, training was always done with the \textit{direct substitution} method because it is the most robust. We show in fig.~\ref{fig:convergence_extended} the results for all combinations of training and inference solve methods. The plots in the first column have no finetuning and therefore are runs with exactly the same parameters. The differences between the plots in the first column are therefore solely based on the random initialization of the NNs.

\end{document}